%% file: main.tex
\documentclass[10pt,twocolumn,letterpaper]{article}

\usepackage[pagenumbers]{cvpr} 

\usepackage{booktabs,threeparttable,multirow,makecell,tabularx}
\usepackage{xcolor}
\usepackage{siunitx}
\usepackage{soul}
\usepackage{colortbl}
\usepackage{color,xcolor}
\usepackage{booktabs}
\usepackage{graphicx}
\usepackage{multirow}
\usepackage{adjustbox}
\usepackage{booktabs}

\input{preamble}
\definecolor{cvprblue}{rgb}{0.21,0.49,0.74}
\usepackage[pagebackref,breaklinks,colorlinks,allcolors=cvprblue]{hyperref}
\usepackage[table]{xcolor}
\usepackage{multirow}
\def\confName{CVPR}
\def\confYear{2026}

\title{Speak While Watching: Unleashing TRUE Real-Time Video Understanding Capability of Multimodal Large Language Models }

\author{%
Junyan Lin$^{1,2*}$ \quad Junlong Tong$^{2,3*}$ \quad Hao Wu$^{2*}$ \quad Jialiang Zhang$^{2,4*}$ \\
Jinming Liu$^{2,3}$ \quad {Xin Jin}$^{2}$ \quad {Xiaoyu Shen}$^{2,\dagger}$\\
$^1$Department of Computing, The Hong Kong Polytechnic University\\
$^2$Ningbo Key Laboratory of Spatial Intelligence and Digital Derivative, Institute of Digital Twin, EIT  \\
$^3$Shanghai Jiao Tong University
$^4$Ocean University of China \\
 \small{
  \href{mailto:email@domain}{junyan.lin@connect.polyu.hk}, \href{mailto:email@domain}{xyshen@eitech.edu.cn}
 }
}

\begin{document}

\newcommand\blfootnote[1]{%
\begingroup
\renewcommand\thefootnote{}\footnote{#1}%
\addtocounter{footnote}{-1}%
\endgroup
}

\maketitle

\input{sec/0_abstract}    
\input{sec/1_intro}
\input{sec/2_related}

\input{sec/3_method}

\input{sec/4_experiment}

{
    \small
    \bibliographystyle{ieeenat_fullname}
    \bibliography{main}
}

\input{sec/X_suppl}

\end{document}

%% file: sec/0_abstract.tex
\begin{abstract}
Multimodal Large Language Models (MLLMs) have achieved strong performance across many tasks, yet most systems remain limited to offline inference, requiring complete inputs before generating outputs. Recent streaming methods reduce latency by interleaving perception and generation, but still enforce a sequential perception–generation cycle, limiting real-time interaction. In this work, we target a fundamental bottleneck that arises when extending MLLMs to real-time video understanding: \textbf{the global positional continuity constraint} imposed by standard positional encoding schemes. While natural in offline inference, this constraint tightly couples perception and generation, preventing effective input–output parallelism. To address this limitation, we propose a parallel streaming framework that relaxes positional continuity through three designs: Overlapped, Group-Decoupled, and Gap-Isolated. 
These designs enable simultaneous perception and generation, allowing the model to process incoming inputs while producing responses in real time. Extensive experiments reveal that Group-Decoupled achieves the best efficiency–performance balance, maintaining high fluency and accuracy while significantly reducing latency. We further show that the proposed framework yields up to 2× acceleration under balanced perception–generation workloads, establishing a principled pathway toward speak-while-watching real-time systems.  We make all our code publicly available: \url{https://github.com/EIT-NLP/Speak-While-Watching}.

\end{abstract}

%% file: sec/1_intro.tex
\section{Introduction}
\label{sec:intro}

\blfootnote{$*$ Equal contribution. $^\dagger$ Corresponding authors.}

Modern Multimodal Large Language Models (MLLMs) have demonstrated remarkable capabilities in a wide range of tasks~\cite{li2024llava,bai2025qwen2,zhu2025internvl3}. However, the vast majority of existing systems still operate under an \emph{offline} inference paradigm, in which the model must first ingest the entire input sequence before producing any output. While this design aligns well with current benchmark settings~\cite{fu2025video,fang2024mmbench,ning2023video}, it inherently precludes real-time understanding and response. 
In practical, safety- and time-critical scenarios, such as assistive navigation~\cite{kuriakose2023deepnavi}, sign language interpretation~\cite{camgoz2020sign}, and live video description~\cite{chen2025livecc}, continuous perceptual feedback is essential. Systems that rely on offline processing cannot react promptly to dynamic changes in the environment, limiting their usability in real-world deployments.

\begin{figure*}[t]
  \raggedright 
  \includegraphics[width=\linewidth]{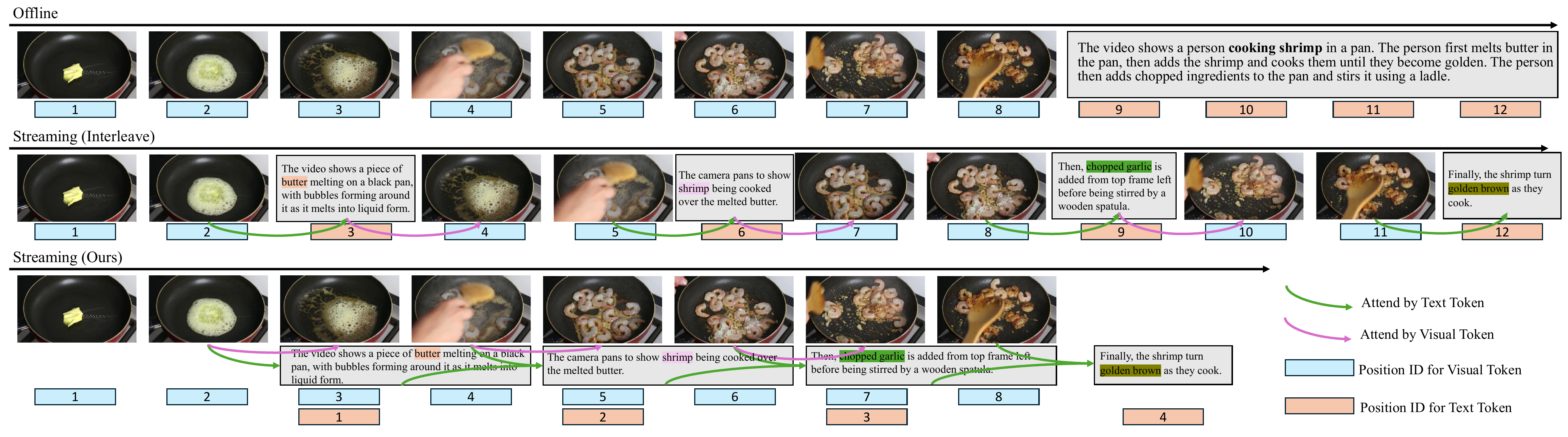}
  \caption{Illustration of different paradigms for video description and positional encoding.The first row shows the offline paradigm, where the model generates the description after observing the entire video, leading to temporal misalignment between narration and visual sequence.
The second row presents the interleaved streaming paradigm, which alternates between perception and generation, providing more immediate responses and better temporal coherence, but still suffers from the continuity constraint of positional encoding that prevents full parallelism.
The third row illustrates our proposed parallel streaming paradigm, which breaks this continuity, enabling simultaneous perception and generation for true real-time video understanding.
The positional IDs serves as a conceptual reference, showing how relaxing positional continuity enables parallel processing between input and output.\textbf{ \textbf{Please zoom in for a clearer view of details.}}}

  \label{fig:fig1}
  \vspace{-10pt}
\end{figure*}

To address this, recent studies \cite{chen2025livecc,qian2024streaming,zhang2024flash} have attempted to extend MLLMs into a streaming paradigm. However, most of these approaches are essentially interleaved: they alternately process a segment of input and then generate a segment of output. Although this reduces latency compared to fully offline inference, it still behaves like ``mini-batch offline'' processing and fails to achieve \textbf{true real-time interaction}. For example, in assistive navigation for the visually impaired, the system may be generating a long descriptive output about the next steps to take. If a sudden obstacle or danger appears during this period, an interleaved streaming MLLM, which performs perception and generation alternately rather than concurrently, may fail to detect the hazard in time. Such behavior is clearly unacceptable in safety-critical applications.
These limitations stem from the fundamental constraints of decoder-only architectures~\cite{dubey2024llama,yang2025qwen3,brown2020language}, which are not designed for simultaneous encoding and decoding. Although encoder–decoder architectures~\cite{raffel2020exploring,lewis2020bart} could, in principle, support such parallelism, converting existing MLLMs to this paradigm is highly impractical, as it would require re-establishing large-scale vision–language alignment from scratch.

We argue that the true bottleneck lies not in the architecture itself, but in the \textbf{positional encoding design}. Current MLLMs enforce a global \emph{continuity} constraint in positional indexing~\cite{su2024roformer,li2024llava,wang2024internvideo2}. Because future output length is unknown at inference time, the model cannot assign consistent positional indices to incoming inputs while decoding is in progress, thereby preventing concurrent perception and generation. As illustrated in Fig. \ref{fig:fig1}, the first row depicts the offline setting for a video description task\cite{bolya2025perception}, where the model generates the full description after observing the entire video. This often results in temporal misalignment between the narrative and the actual video sequence. For example, the model may start describing “shrimp cooking” even though the shrimp does not appear until the middle of the video. The second row illustrates the interleaved streaming setting, where the description follows the temporal order more naturally, first describing the melting of butter, then the cooking of shrimp, and finally the addition of seasonings. Although this result is more temporally coherent, its latency remains suboptimal because the continuity of positional encoding prevents the model from encoding the next incoming frame until the current text generation is completed.

We observe that such strict positional continuity is not fundamentally required. The essential role of positional encoding is to capture \emph{relative relationships} among tokens, rather than to impose a single, globally continuous index space~\cite{tong2025llm,tong2025streamingthinker}. This insight allows us to decouple positional assignments across input and output streams while preserving the relational structure necessary for multimodal alignment. Motivated by this perspective, we introduce a \emph{parallel streaming} paradigm that breaks positional continuity and enables true simultaneous encoding and decoding. As illustrated in the third row of Figure~\ref{fig:fig1}, our approach allows the model to prefill embeddings for incoming visual frames \emph{during} text generation, achieving real-time synchronization between perception and response.

Specifically, we propose three intuitive positional encoding strategies—\textbf{Overlapped Streaming Position Encoding (OSPE)},\textbf{Group-Decoupled Position Encoding (GDPE)}, and \textbf{Gap-Isolated Position Encoding (GIPE)}.
In the OSPE strategy, the model begins encoding the next video frame concurrently with text decoding, assigning the same initial positional indices to both the current response and the next incoming frame.
The GDPE strategy, in contrast, separates the input and output streams, assigning each its own positional group that starts from zero independently.
Finally, the GIPE strategy extends the group-based design by adding a large numerical offset between input and output positions, thus creating an explicit separation in index space.
We conduct extensive experiments under both offline and streaming inference paradigms on Video Description (VD) and Video Question Answering (VQA) tasks, systematically evaluating the proposed positional encoding strategies from the perspectives of performance, robustness, and acceleration potential. Empirically, we observe that in both offline and streaming settings, the original positional embedding scheme can be replaced by our proposed alternatives with only minimal fine-tuning data, while preserving comparable performance across standard evaluation metrics. In terms of robustness, we introduce scheduling perturbations at test time by disrupting the wait-K policy and find that all three proposed strategies consistently yield more fluent and stable language generation than conventional interleaved encoding. Considering both task accuracy and linguistic coherence, Group-Decoupled Position Encoding (GDPE) emerges as the most balanced and effective design. Finally, we provide a theoretical analysis showing that parallel streaming enables up to 2× acceleration under balanced input–output workloads. Importantly, this theoretical speedup is broadly applicable to nearly any streaming MLLM, offering a plug-and-play pathway toward faster and truly real-time inference.

Our contributions are as follows:
\begin{itemize}
    \item We identify the key issue preventing true input–output parallelism in current MLLMs: the unnecessary global \textbf{continuity} constraint of position encoding, and propose a novel and intuitive perspective on positional design.
    \item We introduce three position encoding strategies that enable true parallelism in streaming tasks, allowing simultaneous encoding and decoding without waiting.
    \item We systematically validate the proposed positional encoding strategies under both offline and streaming paradigms, demonstrating that GDPE provides the most effective balance between performance and fluency for real-time streaming.

\end{itemize}

%% file: sec/2_related.tex
\section{Related Works}
\label{sec:related}

\begin{figure*}[h]
  \centering
  \includegraphics[width=0.95\linewidth]{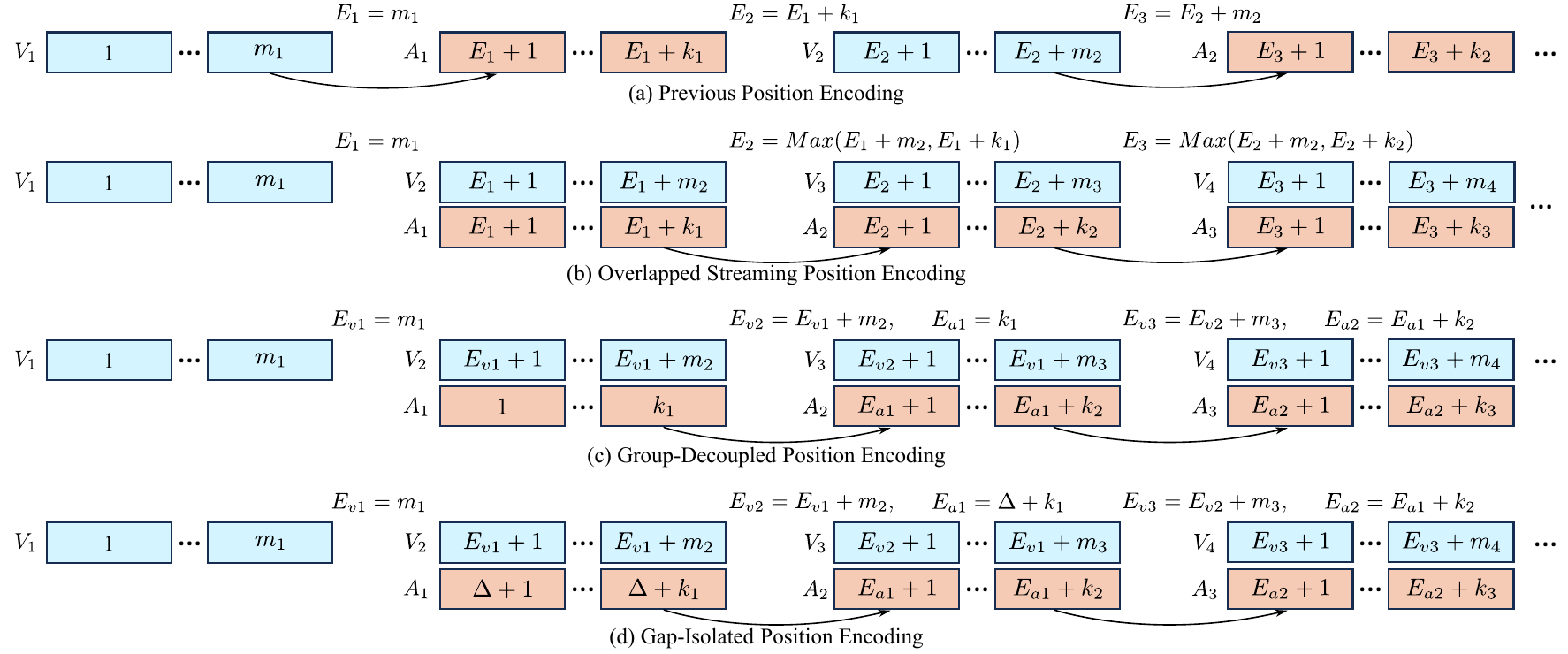}
  \caption{
Comparison of different position encoding strategies, where
$V_i$ represents the video token sequence from the $i$-th input clip, and $A_i$ represents the corresponding textual output token sequence.
Arrows denote the source dependency for the first generated token of each textual output segment.
(a) \textbf{Previous Position Encoding}: assigns consecutive positions strictly following the interleaved video-text streaming order;
(b) \textbf{Overlapped Streaming Position Encoding (OSPE)}: enables video-text streaming parallelism by allowing temporal overlap between encoding and decoding; 
(c) \textbf{Group-Decoupled Position Encoding (GDPE)}: divides video and text into independent groups that maintain intra-group continuity while being inter-group decoupled; 
(d) \textbf{Gap-Isolated Position Encoding (GIPE)}: introduces a fixed gap between groups to fully isolate their index spaces and further reduce cross-modal interference. 
}

  \label{fig:fig2}
  \vspace{-10pt}
\end{figure*}

\paragraph{Streaming Large Language Models}

Most existing multimodal large language models (MLLMs) \cite{liu2023visual,liu2024improved,li2024llava,li2024monkey,ye2024mplug,liu2024sphinx,bai2025qwen2,zhu2025internvl3,tong2024cambrian} follow an offline paradigm, where the model observes the entire video before generating responses. However, this approach faces clear limitations in real-world scenarios. For example, when watching a two-hour movie, users naturally expect interactive, real-time responses rather than delayed answers after viewing the whole video. To address this issue, researchers have begun exploring streaming inference, allowing the model to generate outputs continuously during perception.
Leveraging powerful vision-language pre-training, many studies adopt an interleaved vision-language design to achieve near real-time understanding and generation\cite{zhang2024flash,distreaming,qian2024streaming,chen2025livecc,yao2025timechat,li2025lion,song2024moviechat,xiong2025streaming}. For instance, LiveCC \cite{chen2025livecc} densely interleaves video frames with automatic speech recognition (ASR) transcripts, enabling real-time commentary. 

As the sequence length increases, such interleaved designs suffer from latency accumulation—since prefill and decoding speeds are inversely proportional to the number of tokens—leading to degraded responsiveness in long-video scenarios. Consequently, several recent works have turned to visual token compression and asynchronous perception–generation to improve efficiency.
Flash-VStream \cite{zhang2024flash} introduces a Flash Memory module that enables real-time reasoning over extremely long videos, while TimeChat-Online \cite{yao2025timechat} reduces up to 80\% of visual tokens by exploiting temporal redundancy without breaking positional continuity. ViSpeak \cite{fu2025vispeak} achieves simultaneous input–output by concatenating generated responses with subsequent perceptual inputs, which inevitably mixes heterogeneous semantics within the same embedding space.
In contrast, our method achieves the same goal by redesigning the positional encoding scheme rather than altering the input–output format, thereby preserving the LLM’s intrinsic feature space while still enabling real-time interaction.

\paragraph{Streaming Tasks}
In practical applications, many vision-language tasks naturally operate in a streaming fashion, where input data arrives continuously and the system must respond in real time. For example, live video description \cite{blanco2025live,zhou2024streaming} requires generating descriptive captions for a video stream on the fly, without access to future frames. Similarly, continuous sign language recognition and translation \cite{camgoz2020sign,zuo2024towards} demands interpreting a signer’s continuous video feed into text or speech as it unfolds. In tasks like real-time object tracking \cite{he2018twofold,cao2023observation}, the model needs to continuously localize and describe a target object’s state or trajectory in sequential frames, updating its understanding with each new frame. Another illustrative scenario is interactive streaming video question answering \cite{distreaming,xie2024funqa}, where an agent must answer user queries about a video in real time. In such a setting, a question may be asked before the relevant visual evidence appears, requiring the model to handle temporal asynchrony and retain context until the answer can be given. All these tasks share the characteristic that the input is continuous and time-sensitive. To evaluate the effectiveness of our approach, we conduct experiments on two representative streaming tasks: video description \cite{bolya2025perception} and video question answering (QA) \cite{xie2024funqa}. These tasks are selected for their natural temporal continuity and ease of adaptation to the streaming setting, which allow us to clearly examine the model’s ability to understand partial visual context and produce coherent outputs on the fly.

%% file: sec/3_method.tex
\section{Position Encoding Strategies}
\label{sec:pos-encoding}

\subsection{Limitations of Continuous Position Encoding}

Early models such as the LLaVA series \cite{liu2023visual,liu2024improved} and MiniGPT-4 \cite{zhu2023minigpt} adopt a uniform 1D positional encoding strategy for both visual and textual tokens, following the original design logic of LLMs. While this simplifies training, it overlooks the fact that visual information possesses unique structural dimensions such as height (H), width (W), and temporal axis (T), which differ from text. As a result, recent works increasingly explore 2D or 3D positional encoding strategies (e.g., Qwen2.5-VL \cite{bai2025qwen2}), enabling the model to better understand the spatial and temporal relationships among tokens.
Despite their promising performance, these position encoding strategies all impose a \textbf{global continuity constraint}: every new token must be assigned a position index that strictly follows the used indices. As a result, the position indices of future visual inputs cannot be determined until all previously generated answer tokens have finished decoding. 
Fig.~\ref{fig:fig2}(a) illustrates this position encoding paradigm, where $V_i$ denotes the $i$-th round of visual input with $m_i$ visual tokens, and $A_i$ represents the corresponding textual answer with $k_i$ text tokens. $E_i$ indicates the ending token index of either the visual input or textual output in the $i$-th round. It can be observed that the indexing of the next input or output depends on knowing the length of $m_i$ or $k_i$ from the previous round.
This creates a hard coupling between prefilling and decoding, forcing the model to alternate between input and output in a strictly sequential manner rather than processing them in parallel.

In summary, continuity in position encoding is the \textit{primary obstacle} preventing streaming MLLMs from achieving real-time interaction. To overcome this issue, we revisit the design of position index allocation and propose a unified framework that relaxes global continuity while preserving intra-modal ordering.
We propose three intuitive position encoding strategies:
\textbf{(1) Overlapped Streaming Position Encoding (OSPE)}, 
\textbf{(2) Group-Decoupled Position Encoding (GDPE)}, and
\textbf{(3) Gap-Isolated Position Encoding (GIPE)},
which provide alternative ways to relax global continuity and thereby enable genuine input–output parallelism in streaming environments. For illustrative purposes, we describe our methods using a standard 1D positional indexing scheme as a running example.

\subsection{Overlapped Streaming Position Encoding}

Due to the limitation of position encoding strategies, video segments $V$ and answer tokens $A$ in the previous paradigm are strictly interleaved. 
The most intuitive way to break the continuity is to allow the model to continue ingesting $V_{i+1}$ while generating $A_i$, as if $A_i$ did not occupy additional index space as shown in Fig. \ref{fig:fig2}b. In this case, both $A_i$ and $V_{i+1}$ share the same starting position ID, denoted as $E_i + 1$. The next pair, $A_{i+1}$ and $V_{i+2}$, then start from one greater than the maximum of the end positions of $A_i$ and $V_{i+1}$. In most cases, by the time the model starts generating $A_{i+1}$, $A_i$ has already been completed, since the number of text tokens is usually much smaller than that of visual tokens.

For subsequent rounds, the same rule applies. 
The starting index of both $A_{i+1}$ and $V_{i+2}$ is assigned as one greater than the maximum $E_{i+1}$ of the end indices of $A_i$ and $V_{i+1}$:
\begin{equation}
    E_{i+1} = \max(E_i+m_{i+1},\, E_{i+1}+k_i),
    \label{eq:overlap_update}
\end{equation}
where $E_{i+1} + k_i$ and $E_i + m_{i+1}$ denote the ending indices of $A_i$ and $V_{i+1}$, respectively. 
Here, $k_i$ and $m_{i+1}$ are the numbers of text tokens and visual tokens in the $i$-th and $(i+1)$-th rounds.
This update rule generalizes the OSPE strategy across all rounds, preserving intra-modal ordering while eliminating the global continuity constraint, thereby enabling true parallel streaming.

\subsection{Group-Decoupled Position Encoding}

Fig.~\ref{fig:fig2}(c) illustrates another possible solution, which divides the entire sequence into two independent groups: one for visual inputs and one for textual outputs. Within each group, position indices are assigned continuously, while continuity across groups is removed. This allows new visual inputs to be indexed independently of the textual generation process, effectively decoupling perception and language in the positional space.
In practice, each newly received visual segment $V_{i+1}$ is indexed based only on the end position of the previous visual segment $V_i$, and each newly generated answer $A_{i+1}$ is indexed based only on the end position of the previous answer $A_i$:
\begin{equation}
\begin{split}
    E_{vi+1} &= E_{vi} + m_{i+1}, \\
    E_{ai+1} &= E_{ai} + k_{i+1},
\end{split}
\label{eq:gdpe_update}
\end{equation}
where $m_{i+1}$ and $k_{i+1}$ denote the numbers of visual and text tokens in the $(i\!+\!1)$-th round, respectively.

\begin{figure}[h]
    \centering
    \includegraphics[width=0.45\textwidth]{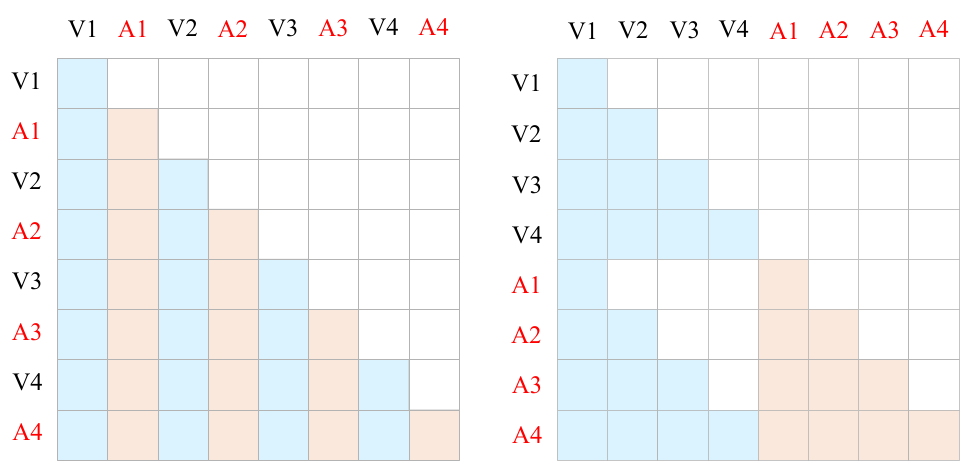}
    \caption{Causal mask visualization. (\textit{left}) Casual mask for previous video-text interleaved streaming paradigm. (\textit{right}) Casual mask for parallel streaming paradigm.}
    \label{fig:causal_mask}
\end{figure}

Although each textual output $A_i$ must still attend to the visual input $V_i$ within the same round, their positional indices reside in separate continuous spaces. This design preserves intra-modal ordering and cross-modal attention while removing inter-modal positional dependency, enabling the model to process visual and textual streams in parallel without violating contextual consistency.

It is worth noting that during training, the input consists of the complete sequences $V_1, V_2, \ldots, V_n$ and $A_1, A_2, \ldots, A_n$, where $n$ denotes all video segments and their corresponding answers. In this process, the causal mask must be carefully set: $V_{i+1}$ should only attend to $V_1$ through $V_i$, while $A_i$ should only attend to $V_1$ through $V_i$ and $A_1$ through $A_i$. A visualization of the causal mask is shown in Figure~\ref{fig:causal_mask}.

\subsection{Gap-Isolated Position Encoding}
While the Group-Decoupled Position Encoding (GDPE) removes cross-modal continuity by assigning independent index spaces to visual and textual groups, their index ranges still remain numerically adjacent within the same overall space. Although it is uncertain whether this adjacency introduces any undesired coupling, we propose \textbf{Gap-Isolated Position Encoding (GIPE)} as a more isolated design that inserts a fixed offset between the two index spaces.
Formally, after assigning indices to all visual tokens $V_1, \ldots, V_n$, the starting index of the first textual token $A_1$ is $\Delta + 1$, where $\Delta$ is a constant gap that isolates the two groups in the positional domain. This ensures that all textual tokens occupy an index range strictly separated from that of visual tokens, making the two modalities positionally disjoint. The causal mask configuration of GIPE remains identical to that of GDPE.

%% file: sec/4_experiment.tex
\section{Experiment}

\subsection{Overview}
We conduct a comprehensive evaluation of our three continuity-breaking position encoding strategies—OSPE, GDPE, and GIPE—built upon the representative 3D spatio-temporal encoding used in recent state-of-the-art MLLMs such as Qwen2.5-VL \cite{bai2025qwen2}. 
Experiments are performed on two tasks, video description and video question answering, to examine how different positional designs affect real-time multimodal understanding.

\subsection{Tasks}

\definecolor{bestcol}{RGB}{7,120,67}   
\definecolor{secondcol}{RGB}{0,102,204}
\newcommand{\best}[1]{\textbf{\textcolor{bestcol}{#1}}}
\newcommand{\secondbest}[1]{\textbf{\textcolor{secondcol}{#1}}}

\begin{table*}[ht]
\centering
\small
\newcommand{\na}{\textemdash}
  \caption{\textbf{Video Description (VD) and VQA on Qwen2.5-VL}.
  Metrics: CIDEr, BLEU-1, BLEU-4, METEOR, ROUGE-L, BLEURT, and Fluency (higher is better).
  \textit{VD} denotes video description task and \textit{VQA} denotes video QA task. For the video QA task, we evaluate the model across all six subsets, but report only the average performance here. Detailed per-subset results are provided in the Appendix.}
\label{tab:vd_vqa}
\resizebox{0.98\textwidth}{!}{
\begin{tabular}{c|c|cccccc|c}
\toprule
{Category} & {Method} & {CIDEr} & {BLEU-1} & {BLEU-4} & {METEOR} & {ROUGE-L} & {BLEURT} & {Fluency} \\
\midrule
\rowcolor{gray!20}
\multicolumn{9}{l}{\textit{Video Description (VD) task}} \\
\midrule
Offline   &  Origin    & 35.44 & 42.36 & 14.45 & 29.18 & 30.47 & 53.21 & 4.84 \\
          &  GDPE      & 30.86 & 40.26 & 13.64 & 28.49 & 34.12 & 53.19 & 4.93 \\
\midrule
Streaming& Interleave & 20.08 & 44.40 & 14.41 & 27.17 & 34.95 & 44.11 & 2.84 \\
          &  OSPE      & 26.32 & 42.14 & 12.78 & 27.92 & 32.29 & 50.62 & 4.48 \\
          &  GDPE      & 12.52 & 26.32 &  7.42 & 30.03 & 27.37 & 51.53 & 4.56 \\
          &  GIPE      & 28.11 & 40.42 & 11.52 & 29.13 & 30.69 & 51.20 & 4.85 \\
\midrule
\rowcolor{gray!20}
\multicolumn{9}{l}{\textit{Video QA (VQA) task}} \\
\midrule
Offline   &Origin    &  6.98 & 34.47 &  4.74 & 19.67 & 22.43 & 41.34 & 4.70 \\
          &GDPE      &  7.25 & 35.23 &  5.13 & 19.50 & 22.98 & 42.04 & 4.52 \\
\midrule
Streaming &Interleave&  3.00 & 21.95 &  1.71 & 13.03 & 18.00 & 31.22 & 3.72 \\
          &OSPE      &  4.22 & 33.40 &  3.68 & 20.23 & 19.61 & 37.38 & 3.98 \\
          &GDPE      &  3.22 & 31.32 &  3.32 & 21.82 & 18.96 & 41.16 & 4.13 \\
          &GIPE      &  3.99 & 30.95 &  2.48 & 17.58 & 19.33 & 37.25 & 4.61 \\
\bottomrule
\end{tabular}
}
\end{table*}

\paragraph{(a) Streaming Video Description.}
In the streaming scenario, the Video Description task aims to generate natural language descriptions for continuously incoming video streams.
Unlike traditional offline captioning, the model must comprehend partial visual context and produce temporally coherent captions on the fly, reflecting real-world applications such as live narration and visual assistance, where minimizing perceptual delay is essential.
We adapt the \textit{PE Video Dataset}\cite{bolya2025perception}, which was originally developed for offline video perception.
The PE Video contains high-quality videos with rich motion dynamics and human-refined captions, making it suitable for streaming scenarios.

\paragraph{(b) Streaming Video QA.}
In the streaming setting, the Video QA task requires the model to answer questions based on continuously arriving video frames rather than full offline clips. The model must reason over partial and evolving visual context, making timely evidence integration essential. We adapt the \textit{FunQA} dataset \cite{xie2024funqa}, which provides diverse human-annotated videos QA pair.
It consists of three subsets: \textit{HumorQA}, \textit{CreativeQA}, and \textit{MagicQA}.
For each subset, we evaluate two task types: video description Q\&A and counterintuitive reasoning Q\&A.
This results in six distinct streaming Video QA sub-tasks, allowing us to comprehensively assess the model’s ability to perform diverse reasoning under streaming conditions.

\subsection{Metric}
For both PE-Video and FunQA tasks, we follow the standard evaluation metrics widely adopted in video captioning and question-answering, including CIDEr \cite{vedantam2015cider}, BLEU \cite{papineni2002bleu}, METEOR \cite{banerjee2005meteor}, and ROUGE \cite{lin2004rouge}.
To better capture the semantic fidelity between generated and reference texts, we further include BLEURT \cite{sellam2020bleurt} as a sentence-level quality metric, which measures contextual similarity beyond surface n-gram overlap.
However, these automatic metrics still fail to reflect the human-perceived fluency and readability of streaming outputs.
Therefore, we additionally employ an LLM-as-Judge evaluation \cite{li2024llms}, where GPT-5 \cite{openai2025systemcard} assesses each generated sentence from a human-like perspective.
Specifically, the model rates linguistic fluency on a 1–5 scale, with higher scores indicating more natural, coherent, and well-structured expressions. The detailed prompt design is provided in the supplementary material.

\subsection{Baseline and Experimental Setup}
We adopt \textit{Qwen2.5-VL} as the baseline in our experiments, which employs explicit three-dimensional positional encoding $(x, y, t)$ for visual tokens, enabling the model to perceive both spatial structures and temporal dynamics.
For textual tokens, the three positional dimensions are kept identical, ensuring consistent positional representation across modalities.
This 3D positional design allows the model to jointly reason over spatial, temporal, and semantic contexts within a unified embedding space.

We adopt a \textit{streaming} evaluation setting based on a fixed \textit{wait-$K$} policy: at test time the model consumes one frame and emits exactly $K=3$ tokens, matching the average frame–token ratio ($\approx 3$) observed in PE-Video and FunQA. Unless otherwise specified, all models are trained and evaluated under this default \textit{wait-$K=3$} configuration.
To ensure a fair comparison, all streaming variants share identical data, optimization settings, and temporal pacing.

Following the sampling protocol of \textit{Qwen2.5-VL}, we set the frame rate to 2 fps. Videos shorter than 5 seconds or longer than 30 seconds are removed.
For each sample from PE-Video or FunQA, we compute the number of text tokens $L$ in its caption/answer and divide it by the video duration $T_{\text{vid}}$ to obtain the average tokens per second.
Let $M = T_{\text{vid}} \times K$ denote the expected caption length under the \textit{wait-$K$} setting.
We discard samples where the response length $L$ is smaller than $M$ (insufficient supervision) or more than twice $M$, since extremely long captions lead to most tokens being emitted at the final frame, causing the generation to behave like offline rather than streaming.
Finally, we randomly select 20K samples for training.
\textit{More details such as dataset examples and additional experimental results are included in the supplementary material.}

\subsection{Performance Analysis}

Table~\ref{tab:vd_vqa} summarizes the performance of all methods under \textit{Offline} and \textit{Streaming} settings.  
Within the Offline category, the \textbf{Origin} model, which fine-tunes Qwen2.5-VL using its native positional encoding, achieves strong results across both Video Description and Video QA. The \textbf{Offline-GDPE} variant replaces the original positional encoding with a GDPE-style layout while keeping the decoding process fully offline. Its overall performance remains close to that of Origin, indicating that modifying the positional layout alone does not fundamentally disrupt the pretrained visual–language alignment, and that such changes can be successfully compensated through limited fine-tuning.

In the Streaming category, the \textbf{Interleave} model which using native positional encoding shows a severe degradation in linguistic fluency compared with both Offline variants. This degradation arises because visual frames are inserted inside the ongoing text sequence, forcing the model to alternate between writing a partial sentence and processing new visual tokens. As a result, the next generated words no longer attend directly to the preceding text token but first encounter the inserted visual tokens in the attention path. This fragmentation disrupts sentence continuity and leads to substantial drops in fluency-sensitive metrics such as BLEURT, revealing that the interleaving mechanism compromises the continuity and readability of the generated text.

In contrast, our continuity-breaking strategies overcome this issue by restructuring the attention order between input and output tokens, as illustrated in Fig.~\ref{fig:causal_mask}, ensuring that visual tokens never interrupt the ongoing textual sequence. Among the three, \textbf{OSPE} resumes each textual segment from the maximum position index of the previous stage’s text output and the current stage’s visual input, which yields uninterrupted text segments with non-contiguous position indices. \textbf{GDPE} and \textbf{GIPE} enforce an independent and strictly continuous index space for input tokens and output tokens.

Building upon these properties, despite altering the native positional encoding and inference paradigm, the three continuity-breaking strategies achieve competitive performance across the two tasks. Among them, OSPE produces lower BLEURT and fluency scores than GDPE, which is consistent with its non-contiguous index updates that limit coherence. GIPE, on the other hand, benefits from the clear separation between input and output numerical position, as well as the minimized interaction distance between words, allowing it to reach fluency levels close to those of the offline models. However, its ability to capture key semantic content is slightly weaker than GDPE. Considering linguistic quality, \textbf{GDPE offers the most balanced overall performance and therefore represents the most promising default configuration for future streaming applications.}

\begin{table}[t]
  \centering
  \caption{
    \textbf{BLEURT of video description under scheduling disturbance.}
    Models are trained with fixed wait-$K{=}3$ and evaluated under both fixed wait-$K{=}3$ and test-time \textbf{Random} schedules.
  }
  \label{tab:avg_waitk_random_vertical}
  \begin{tabularx}{\linewidth}{l | c c c c}
    \toprule
    \textbf{Setting} & \textbf{Interleave} & \textbf{OSPE} & \textbf{GDPE} & \textbf{GIPE} \\
    \midrule
    3$\to$3     & 44.11 & 50.62 & 51.53 & 51.20 \\    
    3$\to$Random  & 40.56 & 50.71 & 51.76 & 51.56 \\
    \bottomrule
  \end{tabularx}
\end{table}

\subsection{Robustness under Scheduling Disturbance}
In real streaming scenarios, video frames and user tokens rarely arrive in a perfectly regular pattern. Multiple frames may be buffered together, responses can be delayed, or the emission rate may fluctuate over time. To simulate such irregular behaviors, we train all models with a fixed wait-$K{=}3$ configuration and evaluate them under both the same fixed schedule and a test-time \textbf{Random} schedule, where the number of emitted tokens per step is randomly perturbed. Due to BLEURT’s sensitivity to sentence-level coherence, we rely on it to assess the impact of scheduling disturbance on streaming generation.

\begin{figure}[t]
  \centering
  \includegraphics[width=0.8\linewidth]{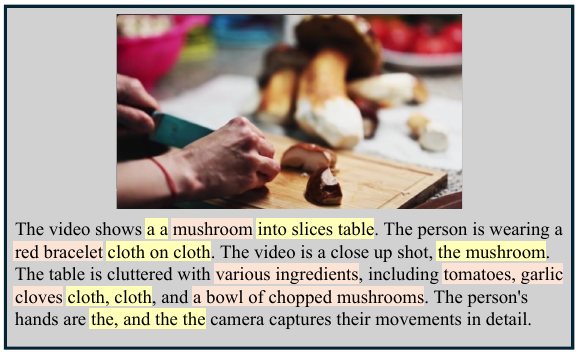}
  \caption{
\textbf{Example of the generated caption by \textit{Interleave} under random scheduling.} 
Duplicated, fragmented, and grammatically broken segments are \textcolor{Goldenrod}{highlighted in yellow}, 
while correctly recognized key objects and actions are \textcolor{red}{highlighted in red}.
}
  \label{fig:case_study_interleave}
\end{figure}

\begin{figure}[t]
  \centering
  \includegraphics[width=0.8\linewidth]{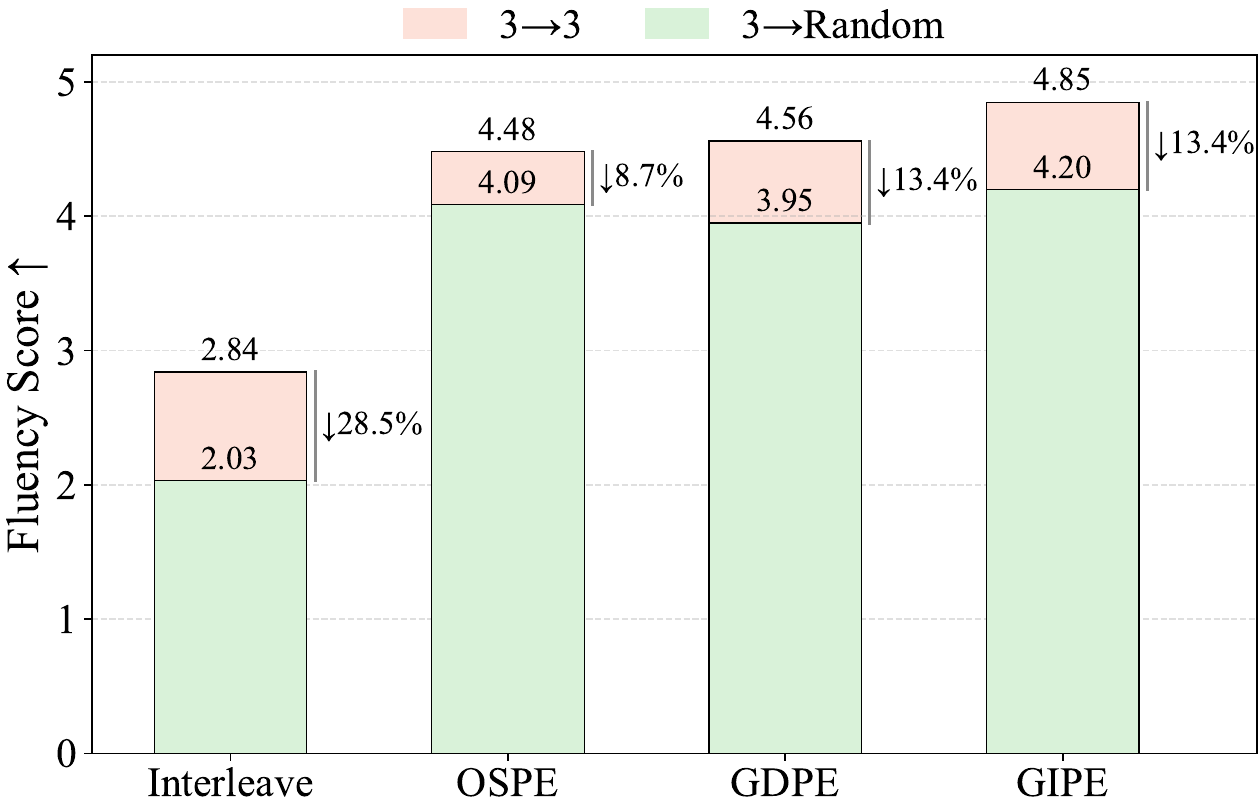}
  \caption{
  \textbf{LLM-as-Judge fluency under scheduling disturbance.}
  The two colors correspond to: 
  (1) trained and evaluated under fixed wait-$K=3$, and 
  (2) trained with wait-$K=3$ but evaluated under random scheduling (disturbance setting).
  }
  \label{fig:fluency_rank}
\end{figure}

The results in Table~\ref{tab:avg_waitk_random_vertical} show that the three continuity-breaking strategies remain stable across settings, whereas Interleave experiences a clear drop under the Random schedule. To further illustrate how scheduling disturbance affects generation, we additionally examine representative outputs together with the fluency evaluation. As shown in Fig.~\ref{fig:case_study_interleave}, Interleave frequently produces duplicated, fragmented, or abruptly truncated phrases when evaluated under random scheduling. These failures arise from the repeated alternation between text generation and visual prefilling, which interrupts sentence progression and causes the model to lose track of its prior context. This qualitative breakdown aligns with the fluency results in Fig.~\ref{fig:fluency_rank}, where Interleave exhibits a pronounced decline under the Random schedule, far larger than that observed for our continuity-breaking strategies. Taken together, these analyses show that Interleave is highly vulnerable to scheduling disturbance, whereas our methods maintain stable and readable outputs even under irregular emission patterns by preserving an uninterrupted textual index space.

\subsection{Theoretical Latency and Speedup Analysis}

In the previous experiments, we have demonstrated that the proposed OSPE, GDPE, and GIPE strategies maintain stable performance under streaming conditions. 
Beyond their accuracy, their core advantage lies in enabling parallel perception and generation by breaking the global positional continuity between input and output tokens, thereby substantially reducing end-to-end latency. 
This subsection further provides a theoretical analysis of the acceleration achieved through such parallelization.

Assume that the entire streaming process consists of $N$ time steps. 
At each step $i$, the model receives $m_i$ visual tokens (perception stage) and generates $k_i$ textual tokens (generation stage). 
Let $R_v$ and $R_t$ denote the visual processing throughput and text decoding throughput (tokens per second), respectively.

\begin{figure}[t]
  \centering
  \includegraphics[width=\linewidth]{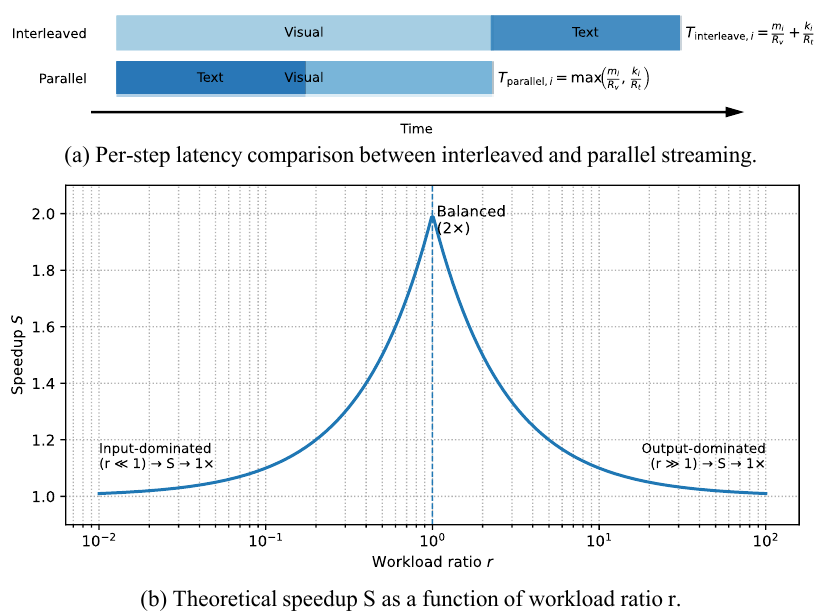}
  \caption{Theoretical latency and speedup analysis.
    (a) Parallel streaming overlaps perception and generation to reduce total step latency.
    (b) The achievable speedup peaks when perception and generation workloads are balanced ($r\!\approx\!1$). \textbf{Please zoom in for a clearer view of details.}
}
  \label{fig:latency}
  \vspace{-10pt}
\end{figure}

\paragraph{Interleaved Streaming (Conventional Paradigm).}
The total latency for the $i$-th step can be expressed as:
\begin{equation}
    T_{\text{interleave}, i} = \frac{m_i}{R_v} + \frac{k_i}{R_t}.
    \label{eq:interleave_step}
\end{equation}
The overall latency across $N$ steps accumulates as:
\begin{equation}
    T_{\text{interleave}} = \sum_{i=1}^{N} 
    \left( 
        \frac{m_i}{R_v} + \frac{k_i}{R_t} 
    \right),
    \label{eq:interleave_total}
\end{equation}
which implies that each stage must wait until the previous one finishes before proceeding to the next, resulting in strictly serialized perception–generation cycles.

\paragraph{Parallel Streaming (Our Paradigm).}
our OSPE, GDPE, and GIPE strategies allow the model to prefetch visual tokens for the $(i{+}1)$-th segment while simultaneously generating textual outputs for the $i$-th step.
Accordingly, the latency per step under ideal parallelization becomes:
\begin{equation}
    T_{\text{parallel}, i} = 
    \max \left( 
        \frac{m_i}{R_v}, \, \frac{k_i}{R_t} 
    \right),
    \label{eq:parallel_step}
\end{equation}
which is evidently smaller than the conventional paradigm. 
In practice, this formulation can be efficiently implemented on two separate GPUs or computational streams, 
where the prefill stage and the decode stage operate in parallel with minimal synchronization overhead.

To further quantify the theoretical acceleration, 
we define the per-step speedup ratio as
\begin{equation}
    S_i = 
    \frac{T_{\text{interleave}, i}}{T_{\text{parallel}, i}}
    =
    \frac{\frac{m_i}{R_v} + \frac{k_i}{R_t}}
         {\max\!\left(\frac{m_i}{R_v},\,\frac{k_i}{R_t}\right)}.
    \label{eq:speedup}
\end{equation}
Let $r = \frac{m_i/R_v}{k_i/R_t}$ denote the workload ratio between perception and generation, 
where $r \gg 1$ indicates vision (input)-dominated latency and $r \ll 1$ corresponds to text (output)-dominated latency.
The relationship between the speedup $S$ and workload ratio $r$ is illustrated in Fig.~\ref{fig:latency}.

This trend can be clearly observed across different tasks.
In \textit{video description} tasks, the model processes long video inputs but generates relatively short textual outputs 
(i.e., $r \gg 1$), resulting in a vision-dominated runtime and only moderate speedup. 
In contrast, \textit{video chain-of-thought (Video-CoT)} involves both extensive perception and long-form reasoning outputs 
($r \!\approx\! 1$), placing it near the balanced regime of Fig.~\ref{fig:latency} and leading to the highest acceleration, 
where the per-step latency is reduced by nearly half compared with the interleaved baseline.

Overall, the achievable speedup is bounded by approximately 2× when perception and generation workloads are balanced, 
whereas the latency asymptotically approaches the perception-only limit as $r$ increases.

\section{Conclusion}
In this work, we revisit the positional encoding design of Multimodal Large Language Models (MLLMs) and reveal that the global positional continuity constraint is the key obstacle to achieving real-time parallel perception and generation.
We propose three continuity-breaking strategies, namely \textbf{Overlapped}, \textbf{Group-Decoupled}, and \textbf{Gap-Isolated} positional encodings, which enable simultaneous input and output without altering the model architecture.
Extensive experiments demonstrate that the Group-Decoupled strategy (GDPE) achieves the best balance between efficiency, temporal coherence, and robustness, significantly reducing response latency while maintaining comparable accuracy to offline models.
Beyond empirical validation, our theoretical analysis confirms that relaxing positional continuity allows genuine “speak-while-watching” capability, achieving up to 2× theoretical acceleration under balanced perception–generation workloads.

\paragraph{Future Work.}
Future research can be explored in the following directions:  
(1) \textbf{Task-specific parallel scheduling}: develop adaptive scheduling strategies tailored to different tasks, enabling the model to dynamically balance performance and latency;
(2) \textbf{Unified streaming framework}: extend the proposed streaming strategies to other modalities such as visual generation, action, and multimodal interaction, forming a unified framework for real-time reasoning;  
(3) \textbf{Hardware-level parallel optimization}: leverage parallel pipelines and multi-GPU execution to further reduce end-to-end latency.  
Through these directions, we view input–output decoupling not merely as a speedup trick, but as a general design principle for future multimodal systems. Extending this idea beyond video to generation, action, and embodied interaction could enable a new generation of MLLMs that reason continuously over the world while speaking, listening, and acting in real time.

%% file: sec/X_suppl.tex
\clearpage
\appendix
\addcontentsline{toc}{section}{Appendix} 
\renewcommand \thepart{} 
\renewcommand \partname{}
\part{\Large{\centerline{Appendix}}}

\section{Overview}
\label{sec:overview}

In this supplementary material, we provide:  
(1) the full prompt used for LLM-as-Judge fluency evaluation \cite{li2024llms,openai2025systemcard};  
(2) concrete input examples from the PE-Video \cite{bolya2025perception} and FunQA \cite{xie2024funqa} datasets under the streaming protocol;  
(3) additional results of the 7B backbone on both tasks and  
(4) funQA sub-task details.

\section{Prompt for Fluency Evaluation}

\begin{figure}[h]
  \centering
  \includegraphics[width=\linewidth]{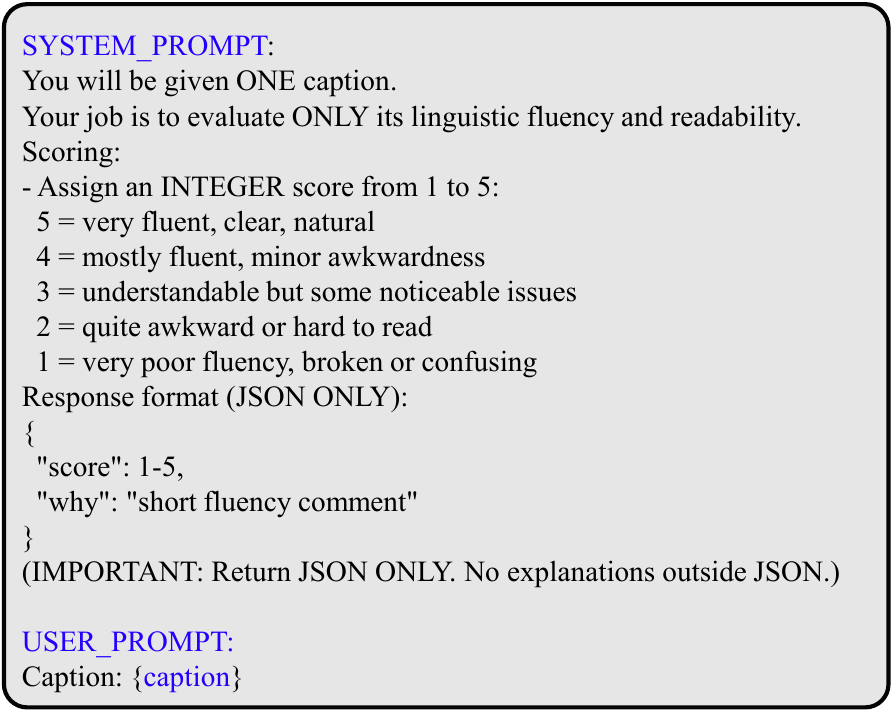}
  \caption{
  Full prompt used for LLM-as-Judge fluency evaluation.
  The judge model receives the task description, the ground-truth caption, and the model output, and then assigns a fluency score from 1 to 5 together with a brief justification.
  }
  \label{fig:sup_prompt}
  \vspace{-8pt}
\end{figure}

Fig.~\ref{fig:sup_prompt} shows the system prompt used for LLM-as-Judge (GPT-5) fluency evaluation. 
The judge receives a single caption and rates \emph{only} its linguistic fluency on a 1–5 scale, returning a JSON dictionary with the score and a short comment. 
This prompt is used across all settings to ensure consistent evaluation.

\begin{figure}[h]
  \centering
  \includegraphics[width=\linewidth]{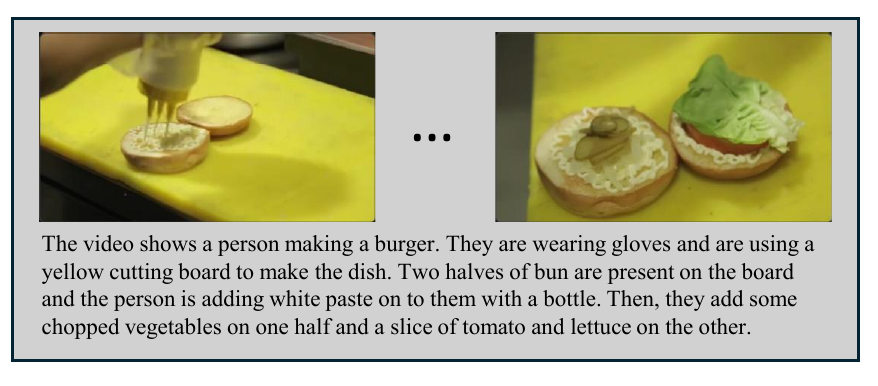}
  \caption{
  PE-Video streaming input example.  
  The model receives frames step-by-step and must produce the caption as the video unfolds.
  }
  \label{fig:sup_pe_input}
  \vspace{-8pt}
\end{figure}

\begin{figure}[h]
  \centering
  \includegraphics[width=\linewidth]{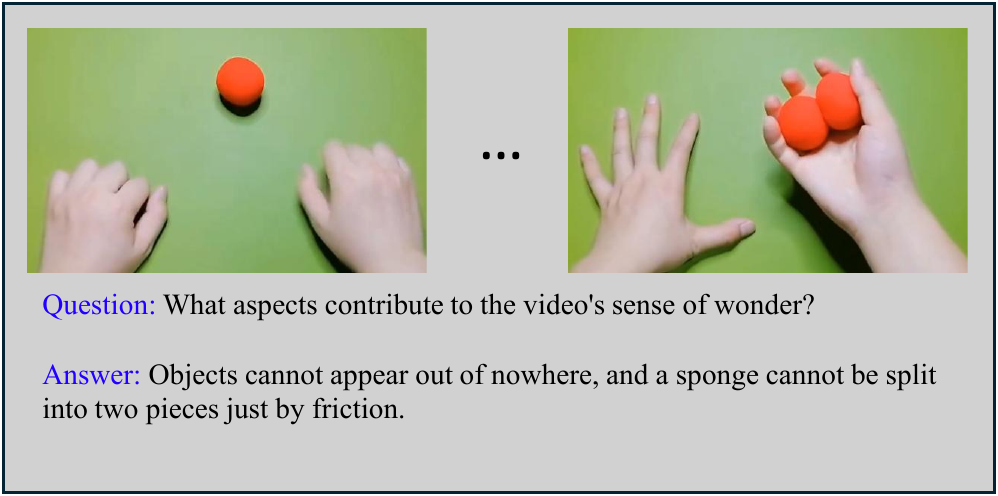}
  \caption{
  FunQA streaming input example.  
  The question is fixed, while the video evidence arrives over time and must be integrated incrementally.
  }
  \label{fig:sup_funqa_input}
  \vspace{-8pt}
\end{figure}

\section{Streaming Input Examples on PE-Video and FunQA}

To better illustrate our streaming protocol, we provide a representative example for each task we test.

\paragraph{PE-Video (Streaming Video Description).}
Fig.~\ref{fig:sup_pe_input} shows a PE-Video example.
The ground-truth captions in this dataset are high-quality and often rely heavily on fine-grained temporal cues, making the task naturally compatible with a streaming formulation where the model must describe the video as frames arrive.

\paragraph{FunQA (Streaming Video QA).}
Fig.~\ref{fig:sup_funqa_input} shows a FunQA sample.
Unlike multiple-choice QA, FunQA requires open-ended, descriptive answers that explain the underlying visual phenomena. 
This makes its output form closely aligned with PE-Video captions, enabling a consistent streaming setup where the model integrates incoming frames to produce a free-form answer.

\begin{table*}[h]
  \centering
  \caption{\textbf{Video Description results on the Qwen2.5-VL backbone (3B and 7B).}}
  \label{tab:vd_waitk3}
  \begin{tabularx}{0.92\linewidth}{l | l | c | c c c c c c }
    \toprule
    Category & Method & Model Size
    & CIDEr & BLEU-1 & BLEU-4
    & METEOR & ROUGE-L & BLEURT \\
    \midrule

    \multirow{4}{*}{Offline}
      & Origin   & 3B & 35.44 & 42.36 & 14.45 & 29.18 & 30.47 & 53.21 \\
      & GDPE     & 3B & 30.86 & 40.26 & 13.64 & 28.49 & 34.12 & 53.19 \\
      \cmidrule(lr){2-9}
      & Origin   & 7B & 42.42 & 40.43 & 12.46 & 27.79 & 32.72 & 53.06 \\
      & GDPE     & 7B & 38.13 & 39.58 & 11.97 & 27.20 & 31.58 & 52.63 \\

    \midrule

    \multirow{8}{*}{Streaming}
      & Interleave & 3B & 20.08 & 44.40 & 14.41 & 27.17 & 34.95 & 44.11 \\
      & OSPE       & 3B & 26.32 & 42.14 & 12.78 & 27.92 & 32.29 & 50.62 \\
      & GDPE       & 3B & 12.52 & 26.32 &  7.42 & 30.03 & 27.37 & 51.53 \\
      & GIPE       & 3B & 28.11 & 40.42 & 11.52 & 29.13 & 30.69 & 51.20 \\
      \cmidrule(lr){2-9}
      & Interleave & 7B & 46.94 & 49.02 & 16.13 & 32.24 & 36.29 & 44.78 \\
      & OSPE       & 7B & 47.49 & 43.85 & 12.10 & 28.05 & 31.86 & 51.71 \\
      & GDPE       & 7B & 37.78 & 41.01 & 11.25 & 27.48 & 30.52 & 51.18 \\
      & GIPE       & 7B & 25.70 & 39.09 &  9.85 & 28.71 & 28.82 & 51.16 \\
    \bottomrule
  \end{tabularx}
\end{table*}

\section{Additional Results of the 7B Backbone on Video Description}

To assess how our positional strategies scale with model capacity, 
Table~\ref{tab:vd_waitk3} presents video description results for the 7B Qwen2.5-VL backbone under both offline and streaming settings.

Scaling the backbone from 3B to 7B yields a pronounced increase in CIDEr, while BLEU, METEOR, ROUGE-L, and BLEURT improve by similar margins across all methods. 
This behavior is expected: CIDEr strongly rewards the recall of salient content words, which larger models capture more reliably, whereas the other metrics remain relatively stable once a reasonable descriptive quality is achieved.  
Crucially, the relative ranking and overall behaviors of all positional strategies remain consistent between 3B and 7B, indicating that our streaming formulations transfer well across model sizes and maintain their effectiveness at larger scales.

\section{FunQA Sub-task Details}
The FunQA dataset contains 12 sub-tasks covering diverse video understanding and reasoning capabilities.
In this work, we focus on the six Description \& Reasoning tasks:
Humor (H2, H3), Creative (C2, C3), and Magic (M2, M3).
In the main paper, we report the average performance across these six sub-tasks to provide a concise and unified summary of the model’s overall behavior.
In this appendix, we further present the detailed per-task results for all six Description \& Reasoning tasks.
All tables in this section follow the same experimental settings as in the main paper (identical wait-K configuration, sampling strategy, and evaluation protocol).
The complete results are provided in Tables~\ref{tab:funqa_m2}–\ref{tab:funqa_c3}.

\begin{table*}[h]
  \centering
  \caption{\textbf{FunQA M2 task performance on Qwen2.5-VL-3B}.}
  \label{tab:funqa_m2}
  \begin{tabularx}{0.85\linewidth}{l | l | c c c c c c }
    \toprule
    Category & Method 
    & CIDEr & BLEU-1 & BLEU-4 & METEOR & ROUGE-L & BLEURT \\
    \midrule

    \multirow{2}{*}{Offline}
      & Origin          
        & 11.48 & 40.54 & 5.93 & 25.23 & 25.70 & 47.33 \\
      & GDPE    
        & 11.75 & 41.30 & 6.46 & 25.54 & 25.93 & 47.47 \\
    \midrule

    \multirow{4}{*}{Streaming}
      & Interleave      
        & 7.48 & 41.62 & 4.04 & 20.06 & 24.03 & 40.59 \\
      & OSPE    
        & 1.96 & 31.06 & 3.79 & 26.55 & 21.61 & 41.16 \\
      & GDPE    
        & 5.77 & 34.66 & 3.19 & 21.92 & 22.12 & 45.47 \\
      & GIPE    
        & 4.48 & 35.02 & 4.62 & 24.38 & 22.48 & 42.04 \\
    \bottomrule
  \end{tabularx}
\end{table*}

\begin{table*}[h]
  \centering
  \caption{\textbf{FunQA M3 task performance on Qwen2.5-VL-3B}.}
  \label{tab:funqa_m3}
  \begin{tabularx}{0.85\linewidth}{l | l | c c c c c c }
    \toprule
    Category & Method 
    & CIDEr & BLEU-1 & BLEU-4 & METEOR & ROUGE-L & BLEURT \\
    \midrule

    \multirow{2}{*}{Offline}
      & Origin          
        & 11.61 & 37.26 & 7.02 & 22.95 & 24.15 & 41.69 \\
      & GDPE    
        & 6.69 & 31.46 & 2.75 & 19.61 & 20.06 & 41.85 \\
    \midrule

    \multirow{4}{*}{Streaming}
      & Interleave      
        & 4.01 & 24.94 & 1.67 & 15.97 & 16.89 & 33.66 \\
      & OSPE    
        & 0.27 & 16.85 & 1.07 & 19.55 & 13.19 & 35.82 \\
      & GDPE    
        & 2.86 & 21.16 & 0.82 & 17.28 & 15.18 & 40.43 \\
      & GIPE    
        & 2.47 & 23.22 & 1.85 & 19.74 & 16.24 & 34.04 \\
    \bottomrule
  \end{tabularx}
\end{table*}

\begin{table*}[h]
  \centering
  \caption{\textbf{FunQA H2 task performance on Qwen2.5-VL-3B}.}
  \label{tab:funqa_h2}
  \begin{tabularx}{0.85\linewidth}{l | l | c c c c c c }
    \toprule
    Category & Method 
    & CIDEr & BLEU-1 & BLEU-4 & METEOR & ROUGE-L & BLEURT \\
    \midrule

    \multirow{2}{*}{Offline}
      & Origin          
        & 10.26 & 38.40 & 4.80 & 21.07 & 22.91 & 39.12 \\
      & GDPE    
        & 13.04 & 40.29 & 5.68 & 20.95 & 23.41 & 39.66 \\
    \midrule

    \multirow{4}{*}{Streaming}
      & Interleave      
        & 4.18 & 17.30 & 1.62 & 9.04 & 13.12 & 26.68 \\
      & OSPE    
        & 3.60 & 30.40 & 3.10 & 22.62 & 19.90 & 36.21 \\
      & GDPE    
        & 8.81 & 37.61 & 4.71 & 20.74 & 22.60 & 39.77 \\
      & GIPE    
        & 6.94 & 35.09 & 3.82 & 20.81 & 21.59 & 39.22 \\
    \bottomrule
  \end{tabularx}
\end{table*}

\begin{table*}[h]
  \centering
  \caption{\textbf{FunQA H3 task performance on Qwen2.5-VL-3B}.}
  \label{tab:funqa_h3}
  \begin{tabularx}{0.85\linewidth}{l | l | c c c c c c }
    \toprule
    Category & Method 
    & CIDEr & BLEU-1 & BLEU-4 & METEOR & ROUGE-L & BLEURT \\
    \midrule

    \multirow{2}{*}{Offline}
      & Origin          
        & 4.71 & 36.52 & 4.08 & 17.45 & 20.55 & 39.80 \\
      & GDPE    
        & 5.03 & 36.39 & 3.69 & 16.55 & 20.54 & 41.64 \\
    \midrule

    \multirow{4}{*}{Streaming}
      & Interleave      
        & 2.09 & 14.00 & 0.76 & 8.11 & 12.33 & 24.88 \\
      & OSPE    
        & 1.94 & 27.15 & 1.66 & 19.70 & 16.85 & 38.02 \\
      & GDPE    
        & 3.63 & 32.22 & 1.77 & 15.85 & 17.09 & 41.57 \\
      & GIPE    
        & 3.42 & 31.55 & 2.26 & 17.97 & 18.59 & 35.47 \\
    \bottomrule
  \end{tabularx}
\end{table*}

\begin{table*}[h]
  \centering
  \caption{\textbf{FunQA C2 task performance on Qwen2.5-VL-3B}.}
  \label{tab:funqa_c2}
  \begin{tabularx}{0.85\linewidth}{l | l | c c c c c c }
    \toprule
    Category & Method 
    & CIDEr & BLEU-1 & BLEU-4 & METEOR & ROUGE-L & BLEURT \\
    \midrule

    \multirow{2}{*}{Offline}
      & Origin          
        & 2.14 & 22.16 & 2.24 & 15.17 & 20.68 & 34.97 \\
      & GDPE    
        & 3.38 & 28.65 & 5.91 & 18.13 & 24.56 & 36.17 \\
    \midrule

    \multirow{4}{*}{Streaming}
      & Interleave      
        & 0.21 & 12.57 & 0.46 & 11.46 & 18.70 & 28.41 \\
      & OSPE    
        & 8.95 & 45.88 & 8.61 & 22.71 & 22.94 & 33.01 \\
      & GDPE    
        & 0.14 & 28.59 & 2.76 & 15.56 & 20.90 & 35.34 \\
      & GIPE    
        & 5.74 & 36.92 & 5.19 & 20.25 & 20.88 & 33.58 \\
    \bottomrule
  \end{tabularx}
\end{table*}

\begin{table*}[t]
  \centering
  \caption{\textbf{FunQA C3 task performance on Qwen2.5-VL-3B}.}
  \label{tab:funqa_c3}
  \begin{tabularx}{0.85\linewidth}{l | l | c c c c c c }
    \toprule
    Category & Method 
    & CIDEr & BLEU-1 & BLEU-4 & METEOR & ROUGE-L & BLEURT \\
    \midrule

    \multirow{2}{*}{Offline}
      & Origin          
        & 1.66 & 31.98 & 4.39 & 16.17 & 20.60 & 39.05 \\
      & GDPE    
        & 3.58 & 33.31 & 6.28 & 16.23 & 23.40 & 39.62 \\
    \midrule

    \multirow{4}{*}{Streaming}
      & Interleave      
        & 0.02 & 21.29 & 1.70 & 13.56 & 22.95 & 30.99 \\
      & OSPE    
        & 2.62 & 36.60 & 1.70 & 19.80 & 19.31 & 34.82 \\
      & GDPE    
        & 2.77 & 31.51 & 1.64 & 14.18 & 18.10 & 37.65 \\
      & GIPE    
        & 2.28 & 38.60 & 4.34 & 18.24 & 17.87 & 35.02 \\
    \bottomrule
  \end{tabularx}
\end{table*}